\documentclass[final]{anthology-ch} 

\usepackage{booktabs}
\usepackage{graphicx}
\usepackage{multirow}
\usepackage{placeins}

\title{Serialization and the Character-System: Measuring Narrative Concentration in French Fiction (1800--1940)}

\author[1,2]{Jean Barré}[orcid=0000-0002-1579-0610]
\author[2]{Antoine Bourgois}[orcid=0009-0008-8117-4120]
\author[2]{Thierry Poibeau}[orcid=0000-0003-3669-4051]

\affiliation{1}{Max Planck Institute for Empirical Aesthetics, Frankfurt am Main, Germany}
\affiliation{2}{LaTTiCe Laboratory, CNRS \& ENS-PSL, Paris, France}

\keywords[eng]{computational literary studies, character networks, character-system, narrative concentration, roman-feuilleton, serialization, distant reading, evaluation}

\pubyear{2026}
\pubvolume{2}
\pagestart{1}
\pageend{1}
\paperorder{1}
\conferencename{Proceedings of Computational Humanities Research 2026}
\conferenceeditors{Editor1 Editor2}
\doi{00000/00000}
\customhead{}

\begin{document}

\maketitle
\begin{abstract}
Every novel divides a finite quantity of narrative attention among its characters, and the shape of that division, between a narrative organized around a single figure and one that spreads itself across a crowd, is one of the form's basic structural choices. Whether that choice has a history is far less settled than the abundance of critical commentary on protagonists might suggest. We operationalize Woloch's notion of the \emph{character-system}, the distribution of narrative attention across a novel's cast structured by the tension between ``the one'' and ``the many'', as the inequality (Gini coefficient) of the PageRank distribution over a character network, a measure of structural concentration that we trace across 12{,}011 French novels published between 1800 and 1940, drawn from the \emph{Hors d'usage} corpus of Gallica fiction. Over the long nineteenth century the trajectory is U-shaped, with concentration declining to a trough in the 1850s--1860s and rising again through the turn of the century. The change is small, comparable to the difference between a novel built on two co-equal protagonists and one built on a single protagonist, but it is robust to the character cutoff, to corpus density, to cast size, and to author effects, and detective and adventure fiction turn out to be less concentrated than the average novel of their period. Literary history reads the \emph{roman-feuilleton} either as the novel of a superior hero or as a form of proliferating episodes, and on this structural measure our results support the second reading. Across the age of the feuilleton (1840--1914), authors whose careers were structured by newspaper serialization distribute centrality more evenly than a literary control, net of period, genre, and canon. Because such claims are only as trustworthy as the networks they rest on, we also evaluate the extraction pipeline against manually annotated novels, at the levels of cast and character identity.
\end{abstract}

\section{Introduction} \label{sec:intro}

Since character extraction became a routine part of computational literary studies, it has supported work on character archetypes \cite{bamman_bayesian_2014, Barre_2025}, gender representation \cite{underwood_transformation_2018}, plot dynamics \cite{konle_character_2023}, and character similarity at scale \cite{bourgois_toward_2026}. That narrative attention is distributed unevenly across a cast is a longstanding observation in the theory of the novel, whereas whether that distribution has itself a history, and in which direction, is far less settled. One might expect the novel to concentrate progressively around a small protagonist core, as focalization and the representation of consciousness are refined over the nineteenth century \cite[206--207]{genette_figures_1972}, and one might equally expect the opposite, since Cohn sees the genre return to interiority in cycles, without linear progress \cite[8--9]{cohn_transparent_1988}, and accounts of the novel as a succession of distinct formal ``technologies'' \cite{paige_technologies_2020} give little warrant for a single unidirectional trend.

In Woloch's account \cite{Woloch2003}, a novel allocates a finite quantity of narrative attention, its \emph{character-space}, and the \emph{character-system} is the distribution of that attention across all characters, structured by a constitutive tension between ``the one'' (the protagonist) and ``the many'' (the minor characters subordinated to, or crowded out by, that centre). Read in these terms, the question becomes whether the inequality of the character-system, the degree to which a few characters monopolize narrative centrality, changes over time. We measure this inequality with the Gini coefficient of the PageRank distribution over each novel's character network (Section~\ref{sec:metric-literary}), a value near zero describing an egalitarian ensemble and a value near one a network dominated by a handful of nodes, so that a rising Gini-of-PageRank reads as a stronger weighting of relational centrality toward ``the one''. Since the index captures the uneven distribution of relational centrality across the cast, and leaves aside the compression, interiority, or voice through which that asymmetry is also realized, it is a partial and structural operationalization of Woloch's concept, and we call what it measures \emph{structural concentration} throughout.

If the distribution of narrative attention has a history, the nineteenth century is the obvious place to look for it, as the century in which the conditions of novelistic production changed most, with the arrival of the \emph{roman-feuilleton} in the daily press from 1836 onwards. On what serialization did to the character-system, literary history holds two opposite views (Section~\ref{sec:serial}). For one tradition, running from Gramsci to Eco and Vareille, the serial novel is the form of the \emph{justicier}, a superior figure such as Sue's Rodolphe or Dumas's Monte-Cristo who dominates the plot and to whom readers return day after day. For another, which Eco also records, instalment publication multiplies episodes and secondary intrigues, so that the fictional population splits into several centres of gravity. Taken structurally, the two readings imply opposite effects on concentration, and only a comparison of serial and non-serial production at scale can decide between them. A similar intuition makes the detective novel, organized around its investigator, a natural candidate for concentration, and we test it as well.

Such a comparison needs the novels that once formed the bulk of production and that literary memory has lost. In the spirit of the distant-reading programme that treats the ``great unread'' as the proper object of literary history \cite{cohen_sentimental_1999, moretti_slaughterhouse_2000, moretti_conjectures_2000}, we therefore work with \emph{Hors d'usage}, a curated corpus of French fiction drawn from Gallica \cite{barre_hors_2026}, which lets us ask whether concentration is a property of the canon, of popular production, or of the novel form at large.

Testing such a claim at scale raises an immediate methodological worry, since the networks are themselves the output of an imperfect NLP pipeline and a diachronic trend in concentration could reflect diachronic variation in extraction quality (older orthography, OCR noise, genre-specific naming). We therefore pair the literary investigation with an evaluation of the extraction pipeline against manually annotated novels, asking whether the resulting networks preserve the quantity our argument depends on, the distribution of centrality across the cast.

\paragraph{Contributions.} The paper proceeds in three steps.
\begin{enumerate}
    \item \textbf{A general problem and its operationalization} (Sections~\ref{sec:related}--\ref{sec:metric-literary}), in which we formalize the diachronic concentration of the character-system, grounded in Woloch and Moretti, and measure it by the Gini-of-PageRank.
    \item \textbf{A confidence check on the extracted networks} (Section~\ref{sec:confidence-metric}), with an evaluation at the levels of cast and character identity on twelve novels and a comparison of gold and predicted concentration indices on three fully annotated ones.
    \item \textbf{A diachronic study at corpus scale} (Section~\ref{sec:results}), which traces the trajectory of concentration across the long nineteenth century, estimates the respective contributions of period, genre, and canon, and uses the mode of publication to decide between the two accounts of the serial novel.
\end{enumerate}

\section{Related Work} \label{sec:related}

\subsection{The Character-System}
The idea that narrative distributes attention unequally across its cast has its principal statement in Woloch's notions of \emph{character-space} and \emph{character-system} \cite{Woloch2003}, in which minor figures are flattened or crowded out for a centre to emerge. Formalist and actantial models \cite{Propp1928, greimas_semantique_2015} offer an instructive contrast, since they make hierarchy and function explicit, whereas Woloch's character-system is useful because it does not reduce characters to functions. A terminological caution follows from Mian et al.\ \cite{mian_computational_2026}, who compare several operationalizations of prominence, find that co-occurrence and dialogue centrality agree on the protagonist in only three quarters of cases, and conclude that ``the `one' is not one'' but a small central cluster. We therefore use ``the one'' as shorthand for the concentrated pole of a distribution, without claiming that a single figure occupies it.

\subsection{Counting and Modeling Characters at Scale}

Among the approaches that handle the cast at scale, one tradition proceeds by enumeration, as when Lavocat \cite{lavocat_letude_nodate} proposes the notion of \emph{demographic style}, the manner in which an author populates a fictional world, measured through the distribution of proper-name occurrences and the ratio of named to unnamed figures. Such populations prove patterned and often unintentional, \emph{Le Rouge et le Noir} and \emph{La Chartreuse de Parme} sharing an almost identical population density, a regularity Stendhal can hardly have designed. Her annotation is manual, a limit she points out herself when calling for collaboration with computational researchers. Enumeration nonetheless stops short of hierarchy, and how many figures inhabit a novel tells us little about how narrative attention is shared among them.

Bamman et al.\ \cite{bamman_bayesian_2014} infer character types across fifteen thousand English novels, modelling the words associated with a character as depending on a background distribution, on the character's type, and on the author. Any corpus-level claim about characters must therefore reckon with the possibility that it measures an author's habits and not a property of the form, a point we return to in Appendix~\ref{appdx:robustness}.

\subsection{Character Networks and Their Limits}
In a third tradition, the cast is represented as a graph whose edges record co-occurrence, dialogue, or interaction \cite{Labatut_2019}, and PageRank \cite{page_pagerank_1999} and related centralities have served to identify structurally dominant figures, including in prior work on character networks of French detective fiction \cite{Barre_2026}. Networks supply what enumeration lacks, a graded hierarchy of positions, and they make inequality measurable, Mian et al.\ \cite{mian_computational_2026} reporting Gini coefficients comparable to the ones we compute below. Their work also states the central objection most sharply, namely that network representations construct significance almost entirely through interaction, which sits uneasily with a theory in which prominence is achieved by suppressing others, since a minor character who converses constantly may accrue centrality while remaining subordinated. They respond by adding interiority, narratorial description, and a directed network of who is discussed by whom.

\subsection{Fiction \emph{Hors d'Usage}}
However different, these three traditions are for the most part synchronic and canonical, establishing what a character-system looks like and how it varies between authors and works, without asking whether the distribution of narrative attention has a \emph{history}. Where diachronic ambitions do appear, they generally rest on collections weighted toward works that remained in circulation. Yet the novels that once constituted the bulk of production are those that have fallen out of it, the fiction Cohen \cite{cohen_sentimental_1999} calls \emph{hors d'usage}, out of use and out of literary memory. If the organization of the character-system responded to the conditions under which fiction was produced and sold, that response should show above all in the mass of ordinary production.

To make that production accessible, previous work has curated the \emph{Fictions littéraires de Gallica} collection \cite{langlais_2021_4751204} into the \emph{Hors d'usage} corpus \cite{barre_hors_2026}. Built on the legal-deposit holdings of the Bibliothèque nationale de France, the collection is estimated to cover about 40\% of nineteenth-century French novels \cite{langlais_reconstituer_2019, barre_hors_2026}, and most of its authors are absent from literary histories and anthologies, with a production that peaks in the 1880s, at the height of the \emph{roman-feuilleton}. 

\subsection{The Serial Novel, between the Superman and the Crowd} \label{sec:serial}
Within that mass, the serial novel has received two opposite readings, the first of which makes the \emph{roman-feuilleton} the form of the \emph{justicier}, a superior figure who dominates the plot. Gramsci held that much so-called Nietzschean ``superhumanity'' had its origin and model in Dumas's Monte-Cristo and not in Zarathustra \cite[``Origine popolaresca del superuomo'', Q.~16]{gramsci_letteratura_1950}, Eco \cite{eco_superuomo_1978} followed this ``superman of the masses'' from Sue's Rodolphe to Rocambole and Arsène Lupin, and Vareille \cite[ch.~II.4]{vareille_homme_1989} traces the detective back to the \emph{justicier} of the popular novel, Rodolphe or Monte-Cristo, whose disguises and marginality he inherits. Daily instalments seem to favour such a centre, since a returning figure is the simplest way to bring readers back. The other reading stresses the opposite tendency of the same medium. In the same book, Eco describes \emph{Les Mystères de Paris} as a tree whose trunk is Rodolphe's search and whose branches are the stories of secondary figures, taken up and dropped along the way, a structure that Sue's biographer Bory called centrifugal and saw as typical of a form obliged to revive its readers' attention day after day \cite{eco_superuomo_1978}. Alliet \cite{alliet_strategies_2025}, in an exploratory study of Balzac's character networks, likewise sees the centripetal organization of the novels of the 1830s, built around one focal character, give way in the 1840s to several centres of gravity, as Balzac turns to a dramatic aesthetic and to serial publication. Eco's image holds both readings at once, a trunk for the one and branches for the many, and our measure asks how much centrality the branches take from the trunk, which neither reading has settled beyond a handful of authors.

\section{Operationalizing the Concentration of the Character-System} \label{sec:metric-literary}

\subsection{Character Network Construction}
\label{sec:character_network_construction}

To build the character networks, we rely on the output of the \textsc{Propp-fr} pipeline, a modular adaptation of BookNLP-fr \cite{booknlp_fr_2024} proposed by Bourgois and Poibeau \cite{bourgois-poibeau-2025-elephant}. The pipeline first detects nested character mentions, including proper names, noun phrases, and pronouns, before clustering all spans referring to the same fictional character through neural coreference resolution.

Following coreference and alias normalization, characters are ranked by mention frequency and the top $N$ entities are retained as network vertices (default $N=20$, with robustness to $N\in\{10,20,30,40\}$ reported in Section~\ref{sec:results}). Following Labatut and Bost \cite{Labatut_2019}, two characters are connected whenever they co-occur within the same sentence, and edge weights correspond to the number of sentence-level co-occurrences. Preliminary comparisons indicate that defining co-occurrence by fixed-size token windows does not affect the results. Finally, weighted PageRank \cite{page_pagerank_1999} is computed on each character network, following the protocol described in \cite{Barre_2026}.

\begin{figure}[ht]
  \centering
  \includegraphics[width=\linewidth]{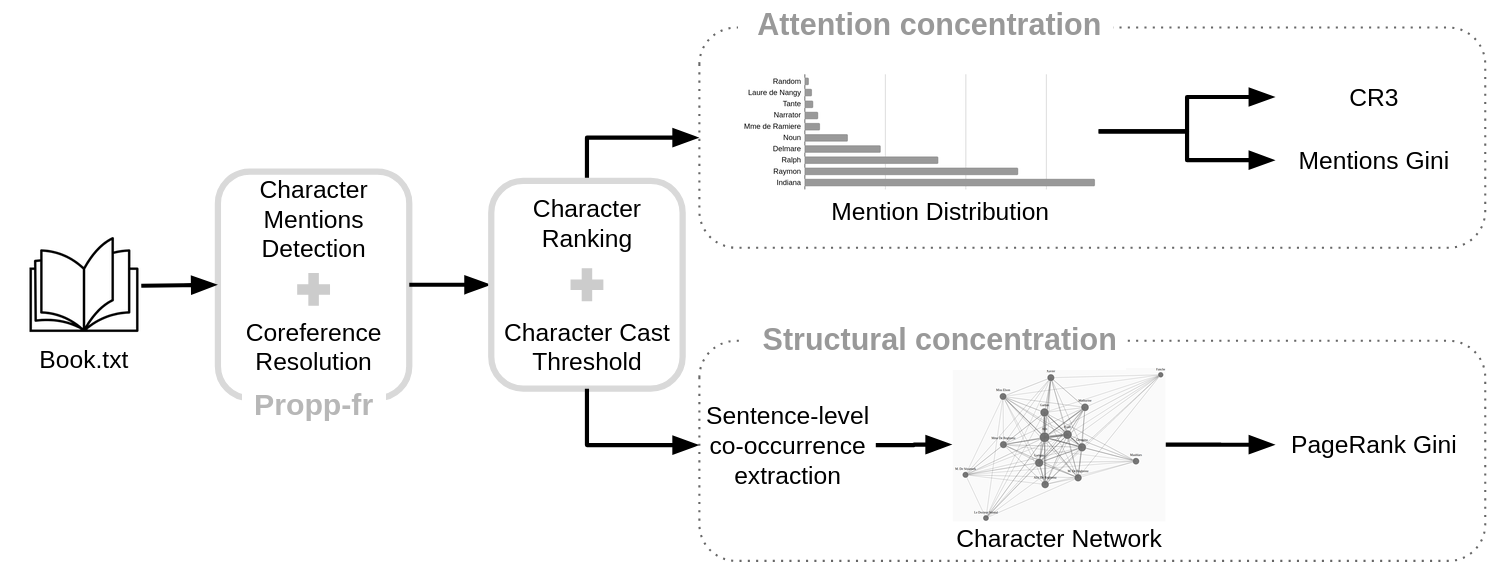}
  \caption{Pipeline Scheme.}
  \label{fig:pipeline_scheme}
\end{figure}

\subsection{Concentration Indices}

We quantify character concentration at two complementary levels, the distribution of textual attention across characters and the distribution of structural importance within the character network. To keep the two apart, we reserve \emph{narrative attention} for the share of mentions a character receives, \emph{structural concentration} for the inequality of PageRank over the network, and \emph{character-system} for Woloch's broader notion, of which both are partial measures.

\subsubsection{Attention concentration}

To measure how strongly narrative attention is focused on a small number of characters, we compute two indices from the distribution of character mentions. The first is the concentration ratio of the three most frequently mentioned characters (CR3, a standard measure borrowed from economics), the proportion of all character mentions accounted for by the three largest characters. The second is the Gini coefficient \cite{ceriani2012origins} of the mention distribution, which comes from the study of income inequality, ranges from 0 (every character equally weighted) to near 1 (one character absorbing the whole distribution), and summarizes how unevenly attention is shared across the cast.

\subsubsection{Structural concentration}

For the network itself, we compute weighted PageRank \cite{page_pagerank_1999} on each novel's co-occurrence graph and derive the Gini coefficient of the resulting distribution. Originally devised to rank web pages, PageRank scores each node by its connection to other well-connected nodes, and thus identifies structurally central characters. The Gini of PageRank is our primary measure of structural concentration, with high values indicating that centrality is concentrated among a small subset of characters. The indices are defined formally in Appendix~\ref{appdx:metric}.

\subsubsection{What the values mean}

Before the measure is put to work, a few novels give a sense of its scale, starting with Féval's serial \emph{Les Habits noirs} (1863), which sits near the bottom of the range at 0.21, built around a criminal association that spreads centrality across a large and mobile population, its leading figure holding only 15\% of all mentions. Flaubert's \emph{Madame Bovary} sits close to the middle at 0.49, Emma being unmistakably the centre while Charles, Rodolphe, Léon and Homais each keep a substantial share of the interactions. Loti's \emph{Pêcheur d'Islande} (1886) sits near the top at 0.61, its small nucleus of figures set against the near-anonymous ground of the fishing community. \emph{Le Père Goriot} (1835), which Woloch reads at length, measures 0.39, on the distributed side, although Alliet reads it as centred on Rastignac \cite{alliet_strategies_2025}, a centrality that grows as the boarding house empties and that a measure computed on the whole novel cannot register.

\section{Validation of Character-Network Extraction} \label{sec:confidence-metric}

\subsection{Evaluation of Character-Network Extraction}

Although character extraction pipelines are usually evaluated through standard coreference resolution metrics such as MUC \cite{vilain1995muc}, B$^3$ \cite{bagga1998algorithms}, CEAF \cite{luo2005coreference}, BLANC \cite{recasens2011blanc}, and LEA \cite{moosavi2016which}, these metrics evaluate the quality of the predicted mention clusters, each emphasizing different aspects of coreference structure \cite{duron2023evaluatecoreference}, whereas our analysis depends on whether the resulting character system preserves the distribution of narrative importance.

A pipeline can therefore achieve reasonable mention-level scores while still introducing the errors that matter here, such as fragmenting an important protagonist, merging distinct characters, or altering the relative centrality of the cast, while errors affecting peripheral characters may leave concentration untouched. Amalvy et al.\ \cite{amalvy2025role} show through perturbation analysis that NER and coreference errors propagate into character networks in ways standard metrics do not anticipate.

\subsection{Reading Three Networks} \label{sec:close-reading}
A first check grounds the measure in individual texts, both to see what it captures when extraction is clean and to expose how the pipeline can distort it. We consider three French novels whose character organization is well understood, \emph{Kernok le pirate} (Sue, 1830), \emph{L'Affaire Lerouge} (Gaboriau, 1866), and \emph{Le Mystère de la chambre jaune} (Leroux, 1908), comparing the pipeline's PageRank ranking of characters against a manually established ranking of narrative importance.

When extraction is clean, the two orderings correspond closely (Spearman $\rho$ between 0.77 and 0.93 once name variants are reconciled), and the measure reflects a character's position in the network and not only how often it is named, so that a figure linking otherwise separate groups can rank above a more frequently mentioned one.

In \emph{Le Mystère de la chambre jaune}, the same comparison brings a failure mode into view, as the coreference stage splits a single character, the antagonist, who appears under several designations, into three distinct chains. His centrality is divided among three nodes, the pipeline ranks him well below his true structural position, and agreement with the reader's ranking drops accordingly ($\rho = 0.79$), while reuniting the chains restores the expected ordering ($\rho = 0.86$). Identity errors of this kind do perturb the concentration measure, but their effect on any single novel is bounded and, in the aggregate, does not overturn the ranking of characters by centrality.

\subsection{Sanity Check on Fully Annotated Books}

We assess whether the complete extraction pipeline preserves the concentration measures used throughout this study by comparing its output against three novels from the extended LitBank-Fr corpus \cite{bourgois-poibeau-2025-elephant}, whose character coreference chains were manually annotated.

\begin{figure}[p]
  \centering
  \includegraphics[width=\linewidth]{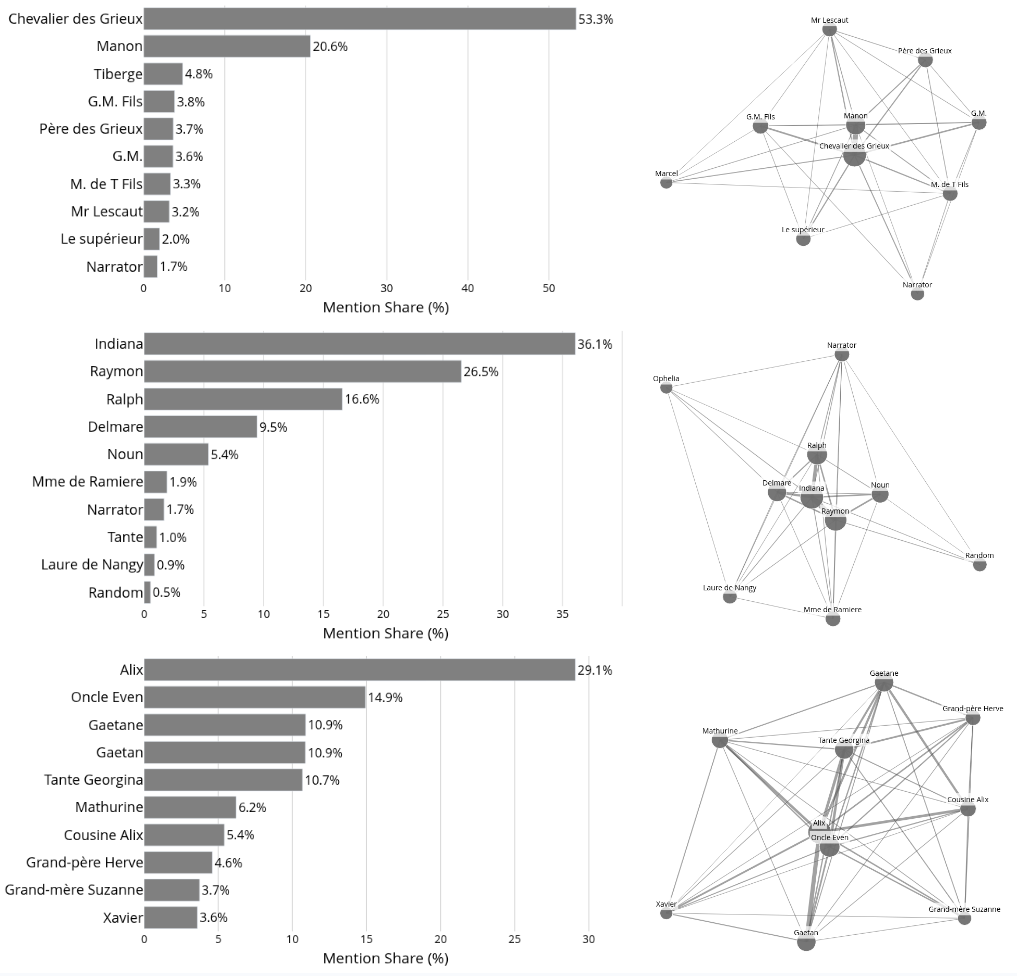}
  \caption{Gold-standard character distributions and co-occurrence networks for the three fully annotated novels. Rows correspond, from top to bottom, to \emph{Manon Lescaut}, \emph{Indiana}, and \emph{Dans les ruines}. The left panels show the mention shares of the ten most frequently mentioned characters, and the right panels the corresponding gold character networks.}
  \label{fig:gold-character-structures}
\end{figure}

For each novel, we construct two networks following the procedure described in Section~\ref{sec:character_network_construction}, one from the gold annotations and one from the pipeline output. We compare the raw character counts before truncation and the concentration indices computed at the operating cutoff $N=20$ (Table~\ref{tab:gold_books_concentration_metrics}), and Figure~\ref{fig:gold-character-structures} shows the gold-standard character distributions and networks of the three novels.

\begin{table}[ht]
  \centering
  \small
\begin{tabular}{lllccr}
\toprule
\textbf{Book} & \textbf{Tokens} & \textbf{Metric} & \textbf{Gold} & \textbf{Predicted} & \textbf{Error (\%)} \\
\midrule
\multirow{4}{*}{\parbox{3cm}{\emph{Manon Lescaut}\\
Abbé Prévost\\
(1731)}}
& \multirow{4}{*}{71,219}
& Character count    & 61  & 106 & +73.8 \\
& & Top-3 character share  & 0.716 & 0.701 & -2.1 \\
& & Mention Gini & 0.723 & 0.725 & +0.3 \\
& & PageRank Gini & 0.564 & 0.545 & -3.4 \\
\midrule
\multirow{4}{*}{\parbox{3cm}{\emph{Indiana}\\
George Sand\\
(1832)}}
& \multirow{4}{*}{112,221}
& Character count    & 81  & 77  & -4.9 \\
& & Top-3 character share  & 0.756 & 0.576 & -23.9 \\
& & Mention Gini & 0.773 & 0.665 & -14.0 \\
& & PageRank Gini & 0.617 & 0.537 & -13.0 \\
\midrule
\multirow{4}{*}{\parbox{3cm}{\emph{Dans les ruines}\\
Delly\\
(1923)}}
& \multirow{4}{*}{95,672}
& Character count    & 59  & 190 & +222 \\
& & Top-3 character share  & 0.486 & 0.504 & +3.8 \\
& & Mention Gini & 0.587 & 0.582 & -0.8 \\
& & PageRank Gini & 0.462 & 0.448 & -3.1 \\
\bottomrule
\end{tabular}
  \caption{Gold and predicted concentration indices for three fully annotated novels at the operating cutoff $N=20$. Values compare the reference annotations (Gold) with model predictions (Predicted), and the error reports relative deviation from gold.}
  \label{tab:gold_books_concentration_metrics}
\end{table}

As Table~\ref{tab:gold_books_concentration_metrics} shows, the raw character count is the least reliable quantity, since the pipeline frequently over- or under-segments coreference chains, leading to substantial differences in the total number of detected characters. This effect is particularly pronounced for \emph{Dans les ruines}, where the predicted network contains more than three times as many nodes as the gold annotation. These discrepancies primarily reflect the difficulty of maintaining coreference chains over long documents, where mentions separated by large textual distances may fail to link to their corresponding entity clusters and produce additional low-frequency characters. This fragmentation can affect major characters as well as peripheral figures.

The concentration indices, by contrast, remain considerably more stable than the raw character count, because they are less sensitive to errors affecting the long tail of the character distribution. For both \emph{Manon Lescaut} and \emph{Dans les ruines}, the Gini coefficient of PageRank differs from the gold value by only 0.019 and 0.014, respectively, despite large errors in character count. The more challenging case, \emph{Indiana}, exhibits larger deviations across all concentration measures, because the pipeline fragments the protagonist into multiple identities, redistributing centrality across several nodes and lowering the apparent concentration of the network. This example shows that errors affecting major characters can meaningfully alter concentration measures.

Even so, the concentration metrics remain substantially closer to the gold values than the raw character counts, which makes them a more reliable basis for corpus-level analysis. In all three novels, however, the predicted Gini of PageRank falls below the gold value, which suggests that extraction errors pull measured concentration downwards. A bias of constant size would leave comparisons between periods or groups of authors intact, so what matters for our argument is whether it varies across them (Section~\ref{sec:failure_modes}).

\subsection{Failure Modes of Character-Network Extraction}\label{sec:failure_modes}

Errors affect network analysis at several levels, since cast errors alter which characters are represented, identity errors fragment or merge nodes, and both can shift the distribution of centrality. Evaluating these errors exhaustively would require manual annotation of all character mentions and coreference links throughout each novel, which is too time-consuming to extend to a larger corpus.

In addition to the three fully annotated novels of the extended LitBank-Fr corpus, we manually compile reference lists of named characters for nine additional novels. These lists identify the main characters using multiple sources and associate each character with a set of unique, unambiguous mentions, each attributable to one and only one character. Three of the twelve novels, \emph{Les Trois Mousquetaires} (1844), \emph{L'Affaire Lerouge} (1866) and \emph{Le Mystère de la chambre jaune} (1908), first appeared as serials in the press, within the age of the \emph{roman-feuilleton} that our analysis singles out.

For all twelve novels, we compute cast and identity evaluation metrics by comparing the pipeline's predictions against the manually established reference data. Table~\ref{failure_modes_table} reports the aggregated results, and the per-novel results are provided in Appendix~\ref{appdx:failure_modes_appendix}.

\begin{table}[ht]
\centering
\begin{tabular*}{0.85\linewidth}{@{\extracolsep{\fill}}lccc}
\toprule
Metric & Mean & Min & Max \\
\midrule
Tokens per novel & 106,981 & 26,127 & 284,497 \\
\midrule
\multicolumn{4}{l}{\textbf{Level 1 -- Cast}}\\
Precision & 100.00 & 100.00 & 100.00 \\
Recall & 89.58 & 62.50 & 100.00 \\
F$_1$ & 93.91 & 76.92 & 100.00 \\
\midrule
\multicolumn{4}{l}{\textbf{Level 2 -- Identity}}\\
Precision & 95.15 & 92.66 & 97.79 \\
Recall & 83.59 & 73.88 & 97.41 \\
F$_1$ & 88.84 & 82.81 & 97.60 \\
\bottomrule
\end{tabular*}
\caption{Cast and identity evaluation across the twelve novels. Precision, recall and F$_1$ are computed per text and averaged.}
\label{failure_modes_table}
\end{table}

Across the twelve novels, identity precision (95.15) exceeds recall (83.59), as it does in every text (Appendix~\ref{appdx:failure_modes_appendix}), which indicates that the pipeline splits characters more often than it merges them, in line with the downward bias noted above. On the three serial novels, identity recall averages 85.5 against 82.9 for the nine others, so on this small sample fragmentation, which would mimic the dispersion we attribute to the feuilleton, does not appear more frequent in serial fiction.

\section{Results} \label{sec:results}

\subsection{Corpora}
\begin{table}[ht]
  \centering
  \begin{tabular}{lrrr}
    \toprule
    Corpus & Novels & Year range & Median year \\
    \midrule
    Hors d'usage (main) & 12{,}011 & 1800--1940 & 1878 \\
    Chapitres (aux.) & 2{,}960 & 1811--2020 & 1855 \\
    \bottomrule
  \end{tabular}
  \caption{Corpora. All analyses reported below are run on \emph{Hors d'usage}. A second corpus, Chapitres \cite{leblond_chapitres_2022}, is used only where an independent point of comparison is informative, and since the two share a majority of their authors, we do not treat them as independent replications.}
  \label{tab:corpora}
\end{table}

The 12{,}011 novels analysed here come from an earlier version of the \emph{Hors d'usage} corpus (Section~\ref{sec:related}) \cite{barre_hors_2026}.
Since catalogue metadata are sparse and unreliable for popular fiction, genre labels come from supervised retrieval, with a classifier trained on a historically situated seed list drawn from the literary history of the genre, applied to the whole archive and refined over a $k$-nearest-neighbour similarity graph \cite{barre-2026-search}. Validated against the independently annotated Chapitres corpus, the procedure recovers 81\% of the novels labelled there as adventure fiction \cite{barre-2026-search, barre_hors_2026}, so what it operationalizes is a situated critical perspective whose boundaries are graded accordingly. Canon status comes from separate work aggregating indicators such as school curricula, prestigious series, and literary prizes \cite{barre_operationalizing_2023}. Novels lacking a label are excluded from the corresponding analyses.

\subsection{The Concentration of the Character-System over Time} \label{sec:res-ushape}
\begin{figure}[ht]
  \centering
  \includegraphics[width=\linewidth]{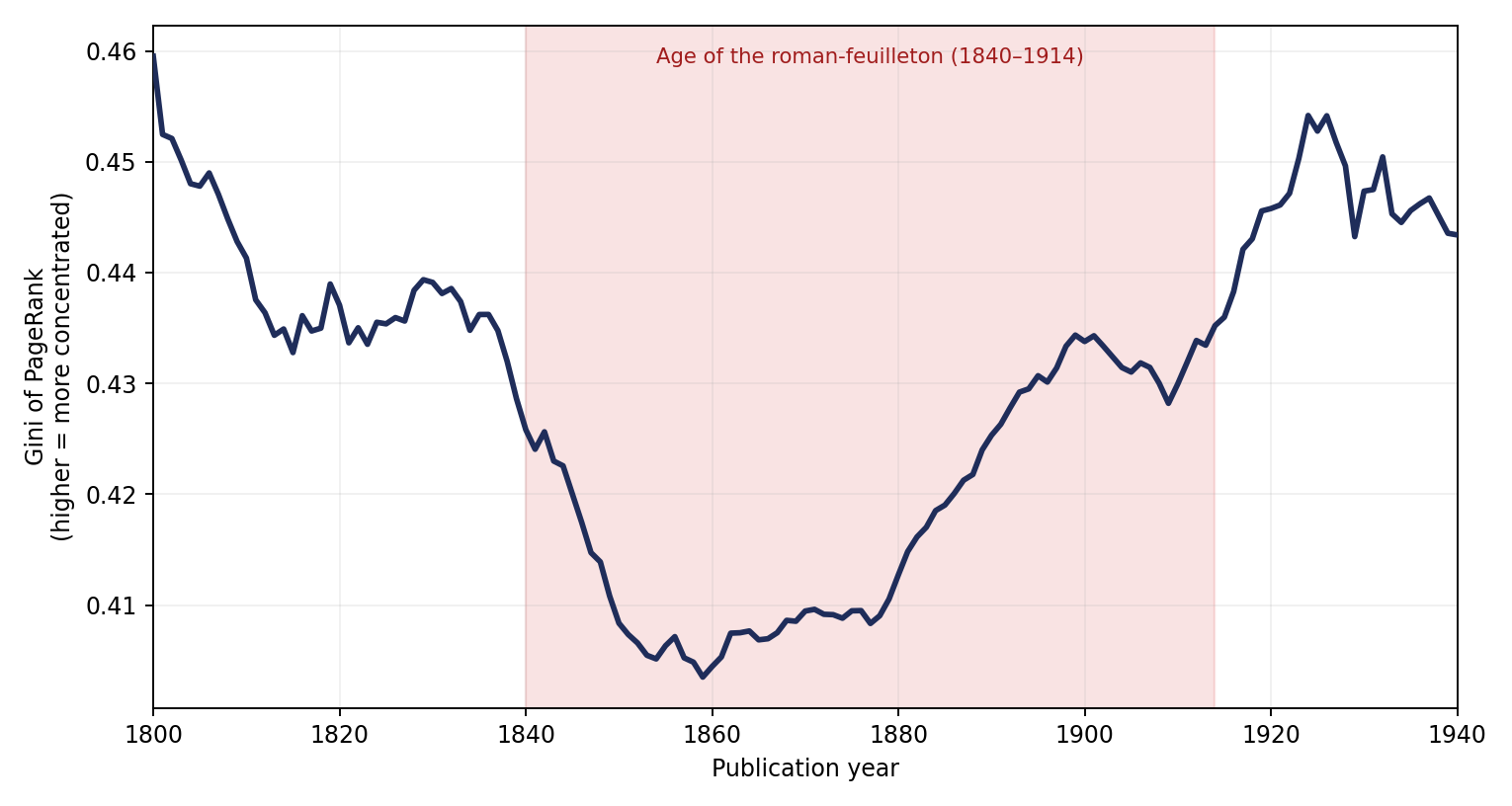}
  \caption{Structural concentration (Gini of PageRank) over time, Gallica ($n=12{,}011$), as a 15-year rolling mean over publication year. The shaded band marks the age of the \emph{roman-feuilleton} (1840--1914), discussed in Section~\ref{sec:res-feuilleton}.}
  \label{fig:ushape}
\end{figure}

Averaged by year across the corpus, structural concentration stands relatively high in the early nineteenth century, declines to a minimum in the 1850s--1860s, and rises again through the turn of the century. A quadratic fit to the yearly means is convex, with its vertex at 1868, so that the overall shape is a shallow U over the long nineteenth century (Figure~\ref{fig:ushape}). Comparing before and after 1900, the rise is statistically unambiguous given the corpus size (a standardized mean difference of about 0.2, $p<10^{-4}$ with errors clustered by author).

Between the trough (1855--1875) and the peak (1920--1940), the mean Gini of PageRank moves from 0.407 to 0.451. For a cast of twenty characters, a shift of that size is roughly what separates a novel organized around two co-equal protagonists from one organized around a single protagonist, the remaining cast held fixed, and it is indeed the leading character that grows, its share of all mentions rising from 0.26 to 0.31 over the same span while the second and third characters hold their ground. The trend thus reads as a gradual tilt of the character-system toward a single centre, with no thinning of the cast.

Set beside the variation between individual novels, the shift is nonetheless small, since the book-to-book standard deviation within a period is about 0.11, some two and a half times the historical change, and roughly a third of the novels at the trough are already more concentrated than the average novel at the peak.

Under the controls reported in full in Appendix~\ref{appdx:robustness}, the trajectory remains essentially unchanged, as varying the character cutoff between ten and forty moves the vertex by only a few years, and because that cutoff holds the number of retained characters nearly constant, cast size cannot account for the curve either. Gallica's production is unevenly spread over time, peaking between roughly 1850 and 1890, yet subsampling a fixed number of novels per year preserves both the shape of the curve and the position of its trough. Author identity explains about a quarter of the variance in concentration, but the U-shape persists, and slightly sharpens, once author fixed effects are included, so the trend does not rest on a handful of prolific writers.

\subsection{Period, Genre, and Canon} \label{sec:res-drivers}
Since publication period, genre, and canonicity are entangled in the corpus, with genres unevenly spread across time and canonized works unevenly distributed, their contributions are best estimated jointly. We therefore regress the concentration measure on all three simultaneously, following the approach used for verb-tense change in French fiction \cite{gabay2025}, with standard errors clustered by author here and in what follows, as several novels may share an author. Period is coded in twenty-year bins, with 1860--1880 as the reference category, since it is both the most densely populated bin (2{,}997 novels), which narrows the confidence intervals of every other period estimate, and the one containing the trough, so that each period coefficient reads as the amount by which concentration stands above its mid-century minimum.

\begin{figure}[ht]
  \centering
  \includegraphics[width=0.92\linewidth]{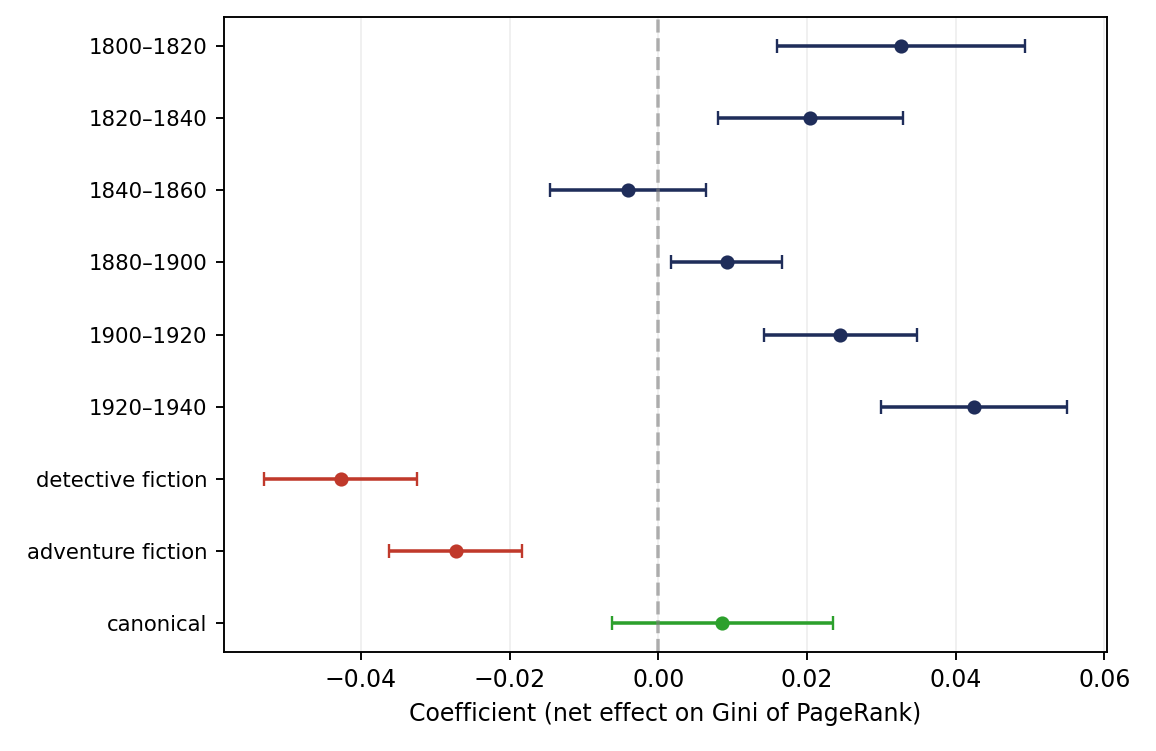}
  \caption{OLS coefficients for the concentration of the character-system, estimated jointly for publication period, genre, and canonical status. Period coefficients (blue) are relative to 1860--1880, genre coefficients (red) are negative, and the canon coefficient (green) is positive. Bars show 95\% confidence intervals.}
  \label{fig:ols}
\end{figure}

Taken together, the estimates qualify any account that would attribute the trajectory to a single genre, since the period coefficients recover the U-shape independently of genre and canon, with concentration higher at both ends of the century than at the trough, whereas the genre coefficients run opposite to what a genre-driven account would predict. Both detective and adventure fiction are associated with lower concentration than the average novel of the same period (detective $-0.043$, adventure $-0.027$, both $p<10^{-4}$ with author-clustered errors), genre fiction keeping its crew and its gallery of suspects without concentrating the narrative on a single figure, even as it participates in the general upward movement. Canonical works show a small association in the other direction ($+0.009$), which does not survive clustering by author ($p=0.26$) and which we therefore do not interpret.


\subsection{Genres and Authors} \label{sec:res-authors}
The Gallica metadata identify only detective and adventure fiction. We extend the comparison using the finer genre classification available in the Chapitres corpus. Ranking its genres by mean concentration over the same period yields a clear gradation (Appendix~\ref{appdx:strata}), with at the low end the forms built on a collective or serially renewed cast, such as short-story collections ($-0.07$ relative to the corpus mean), adventure ($-0.03$), and detective fiction ($-0.01$), and at the high end the forms organized around a personal plot, chiefly the sentimental novel ($+0.08$), the epistolary novel ($+0.07$), and juvenile fiction ($+0.04$). Detective and adventure fiction thus appear as two instances of a broader regularity, in which what distinguishes genres is whether their conventions call for a crew, a gallery, a succession of episodes, or for a single destiny followed through.

\begin{figure}[p]
  \centering
  \includegraphics[
    width=\linewidth,
    height=0.80\textheight,
    keepaspectratio
  ]{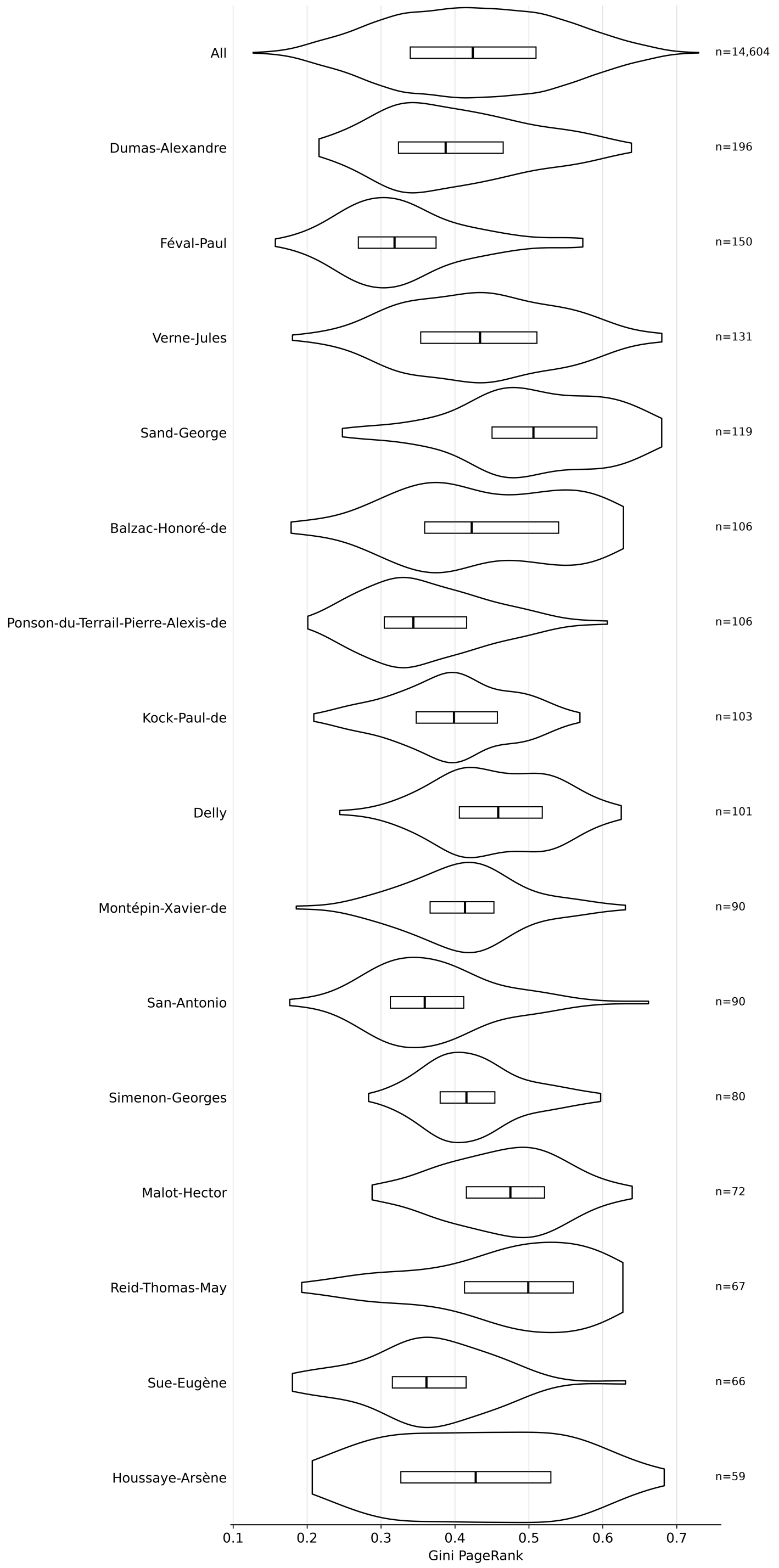}
  \caption{Distribution of the Gini of PageRank for the fifteen most
  represented authors, against the corpus as a whole (merged corpora). Authors differ systematically in how they distribute narrative centrality, and the
  writers of the serial press cluster toward the lower (more distributed)
  end.}
  \label{fig:authors}
\end{figure}

Turning to authors, an author fixed-effect model accounts for between a sixth and a quarter of the variance, depending on how many novels an author must have to enter the model, and individual writers display stable and recognizable profiles (Figure~\ref{fig:authors}). Some organize their novels, book after book, around a small dominant core, as Sand, Delly and Malot do, while others such as Féval, Ponson du Terrail and Sue work consistently with a dispersed cast. Balzac, whom Woloch reads at length, sits on the distributed side, against the intuition that the \emph{Comédie humaine} is organized around its protagonists, although Alliet's observations (Section~\ref{sec:serial}) would lead one to expect his novels of the 1830s to be more concentrated than those of his serial years.

\subsection{Serialization-Oriented Authorship} \label{sec:res-feuilleton}

In light of the two accounts of Section~\ref{sec:serial}, the decades over which concentration reaches its minimum coincide with the age of the \emph{roman-feuilleton}, the novel published in newspapers. Installed in the French press from 1836, the serial became a major form of media culture with Sue's \emph{Les Mystères de Paris} in 1842 and progressively gave way to other supports once the First World War had disorganized the economy of the mass press, so that 1840--1914 marks the media pre-eminence of the serial novel, which never amounted to a domination over all novelistic production \cite{letourneux_fictions_2017, kalifa_culture_2001}. If the proliferation reading is right, novels written for the feuilleton should be measurably less concentrated than their contemporaries published in volume, and the superman reading predicts the reverse.

As our corpus carries no per-work indication of publication venue, we test the hypothesis at the level of the author. Literary history identifies writers whose careers were structured principally by press serialization \cite{queffelec_roman-feuilleton_1989}, among them Sue, Féval, Montépin, Gaboriau, Mary, Zaccone, Lermina, Mérouvel, Belot, and, later in the century, Lavedan, Mendès and d'Ivoi, and, on the other side, writers whose novelistic practice was not principally organized by the popular feuilleton, from Balzac, Stendhal, Sandeau, Murger and the Goncourt to Huysmans, Renard, Régnier and Boylesve. Labels are assigned from publication history alone, never from the measured concentration, and authors who moved substantially between the two circuits (Zola, Sand, Daudet, Maupassant, Verne) are excluded. The control is not free of serial publication, since Balzac, among others, published several of his later novels as serials \cite{alliet_strategies_2025}, which can only narrow the measured gap if serialization does disperse the character-system. Within the window of analysis this yields 1{,}108 novels by 35 feuilleton authors and 794 by 65 authors in the literary control. Because such groupings involve choices that could in principle favour a result, we report the estimate under each of twelve specifications of the two groups (Appendix~\ref{appdx:specs}), where the coefficient ranges from $-0.029$ to $-0.078$, is negative in all of them, and is significant at the 5\% level in ten of the twelve.

\begin{figure}[ht!]
  \centering
  \includegraphics[width=\linewidth]{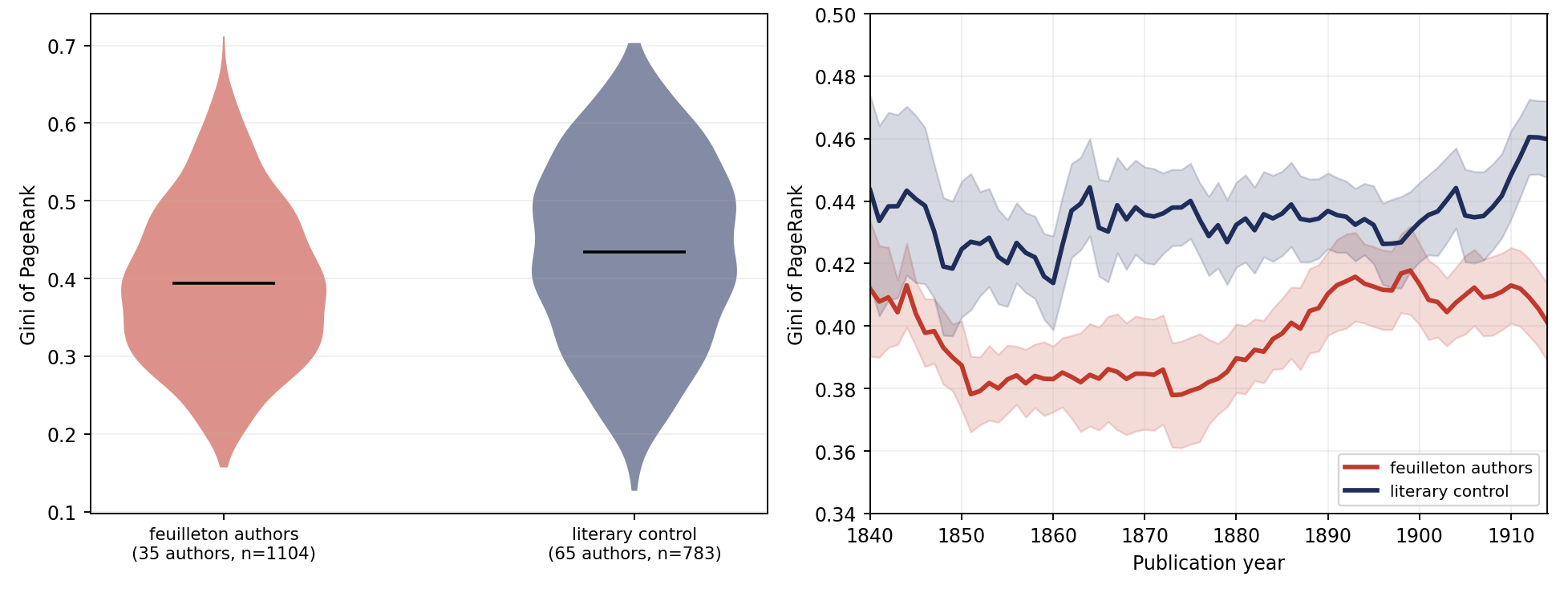}
  \caption{Concentration for feuilleton authors (red) against a literary control (blue). The left panel shows the distributions with group means, and the right panel a 15-year rolling mean with $\pm 1$ standard error, bootstrapped over authors, across the age of the \emph{roman-feuilleton}.}
  \label{fig:feuilleton}
\end{figure}

As Figure~\ref{fig:feuilleton} shows, the feuilleton group has a mean Gini of PageRank of 0.393 against 0.433 for the control. Because the label is a property of the author and not of the individual work, the novels are not independent observations, and we take the author as the unit of analysis throughout. Averaging within writers, the 35 feuilleton authors stand at 0.392 against 0.441 for the 65 authors of the control ($d=-0.84$, $p<10^{-4}$). Since feuilleton authorship is also entangled with period, genre, and canonicity, we estimate the association net of those factors in the regression above, with standard errors clustered by author, which gives a coefficient of $-0.030$ (95\% CI $[-0.053,-0.007]$, $p=0.009$) that grows when the two most publication-ambiguous authors are removed. A permutation test that reassigns the label across authors, holding the number of feuilleton authors fixed and drawing $10^5$ times, returns $p=0.024$. Authors of the serialized novel thus distribute narrative centrality more evenly than their in-volume contemporaries, by an amount that period, genre and canonical status do not absorb and that is comparable to the whole secular movement traced in Section~\ref{sec:res-ushape}, which spans 0.043 from trough to peak.

Since serial novels are well known for their sprawling casts, the result might be mistaken for a restatement of that fact. Our networks, however, are capped at twenty vertices, so what the measure compares is how centrality is shared among the twenty most prominent characters, whatever the size of the full cast. Restricting the comparison to novels that reach the cap exactly, with an identical number of retained characters on either side, leaves the difference intact (author means of 0.391 against 0.438, $d=-0.82$, $p<10^{-4}$), and the leading character of a feuilleton novel holds 24\% of all mentions against 30\% for the control.

Since the gap between the two groups is widest in the middle decades and narrows after about 1880, we turn to the serial group itself over that stretch. Concentration among feuilleton authors is flat before 1880 ($p=0.99$) and rises thereafter ($+0.0014$ per year, $p=0.039$ with errors clustered by author, and $+0.0018$, $p=0.024$, once author fixed effects absorb changes in the composition of the group). The rest of the corpus rises over the same span as well, and a direct test of the group $\times$ year interaction does not separate the two rates ($+0.0011$, $p=0.27$), so we claim only that concentration also rises within the serialization-oriented group. What produces that internal shift is harder to establish, and several natural candidates do not survive testing. Since the cutoff holds the cast fixed, a smaller cast cannot be the cause, and the narrative fabric does not thin out either, as the number of distinct interactions per novel among these authors does not decline after 1880, even though sparser networks are indeed more concentrated ($\rho=-0.39$). A third candidate is the recurring hero, the device most often invoked to explain how serial fiction binds its readers and the one on which the superman reading of the feuilleton relies. Within the predefined group of serialization-oriented authors, those who built series around a returning figure are less concentrated than the remaining authors (7 vs.\ 28 authors, $d=-1.78$, $p<10^{-4}$). Recurring heroes therefore cannot explain the post-1880 rise, although the size of the estimate depends on which series authors one admits (Appendix~\ref{appdx:specs}). What changes after 1880 is thus the internal hierarchy of a network that remains just as populous.
\section{Discussion} \label{sec:discussion}

Of the two readings of the serial novel set out in Section~\ref{sec:serial}, our results favour the second, since novels by authors associated with the serial press have less concentrated character networks than those of the literary control. The association persists after adjustment for period, labelled genre and canonical status, and its magnitude is comparable to the historical change from trough to peak. Within the serialization-oriented group, authors of recurring-hero series also have lower mean concentration, including when the creators of Rocambole and Arsène Lupin, two of Eco's supermen, are included in the comparison (Appendix~\ref{appdx:specs}).

In these novels, the returning hero seems to provide continuity across episodes while secondary plots give other characters substantial roles in the network. His powers and the point of view granted to him can sustain his prominence without requiring him to monopolize interactions. This interpretation preserves the heroic dominance emphasized by the superman reading while accounting for the more even distribution of relational centrality found among serialization-oriented authors. Our measure captures this distribution, leaving aside the forms of prominence through which a hero dominates the fictional world or the reader's attention.

Eco and Bory's observations on Sue suggest that successful secondary plots could be extended over further instalments. Readers' attachment to familiar characters and commercial incentives to prolong publication offer further explanations for the sustained role of supporting casts. Our data cannot distinguish these mechanisms or establish a causal effect of serialization, since publication venue is unavailable at the work level and the association concerns authors' careers, encompassing writing practices, editorial constraints and rhythms of publication.

Structural concentration follows a shallow U over the long nineteenth century, a trajectory that survives the controls reported in Section~\ref{sec:res-ushape} but remains small in amplitude beside the variation between individual novels. Detective and adventure fiction are less concentrated than their contemporaries, despite the prominent heroes associated with both genres. In detective fiction, the inquiry can give suspects, witnesses and victims substantial roles in the network even when the investigator directs the plot. This interpretation accommodates the heroic prominence that Vareille traces back to the \emph{justicier}, although establishing how these roles contribute to the measured distribution requires closer examination of individual novels. Concentration also rises within the serialization-oriented group after 1880, and the causes of this internal change remain unresolved.

We developed the comparison between the two readings of the serial novel after observing the historical trajectory, so it remains exploratory. If instalment publication contributes to lower concentration, novels first published as serials should differ from novels first published in volume by the same authors, within comparable periods and genres. Alliet's account of Balzac offers a case for examining this relationship, and similar comparisons could be made in English and Italian fiction. The comparison with Alliet's reading of \emph{Le Père Goriot} also raises a question about the scale of analysis, since a whole-novel index cannot show how a character becomes central as the plot develops. Networks computed over successive segments would allow us to examine that process and determine whether the same overall concentration conceals different narrative trajectories.

\section{Limitations} \label{sec:limitations}

As a measure, the Gini of PageRank is a partial operationalization of Woloch's character-system, capturing structural asymmetry but not the compression, interiority, or voice through which a novel also realizes the hierarchy between ``the one'' and ``the many''. The effect is modest beside the book-to-book variance and describes the corpus in aggregate, not individual novels. The corpus thins at its temporal extremes, so the robust window is approximately 1830--1930, and it also contains translations into French alongside French-authored fiction, which we do not filter out. It predates the deduplication of the released collection, so some works may be counted in more than one edition.

Genre and canon labels are imperfect, the metadata join is incomplete, notably for pre-1800 and OCR-variant titles, and the concentration metric is conditional on modelling choices ($N$, the sentence window). The test of the serialization hypothesis labels novels by their author and not by their actual mode of first publication, so it supports the hypothesis without establishing it at the level of the individual text. Its effective sample is the 100 authors and not the 1{,}902 novels they wrote, which is why we report author means and cluster the regression errors accordingly. Finally, the evaluation of the pipeline rests on twelve novels, only three of them fully annotated, which is too few to exclude that the downward bias of extraction errors varies with period or mode of publication, a real risk since disguise and multiple identities, which coreference resolution tends to split, are stock devices of the serial hero \cite[ch.~II.4]{vareille_homme_1989}, and extending it across periods, genres and levels of OCR quality is the most pressing methodological task ahead.

\section{Conclusion}

We operationalized the diachronic evolution of Woloch's character-system as the structural concentration of character networks, measured by the Gini of the PageRank distribution, and paired it with an evaluation of the extraction pipeline against manually annotated novels. Tested on twelve thousand French novels, the concentration of narrative centrality proves to have a history over the long nineteenth century, a shallow U with a trough in the 1850s--1860s, which no single genre drives. Where literary history has read the \emph{roman-feuilleton} either as the novel of the superman or as a form of proliferating episodes, our results side with the second reading on this structural measure, authors of the serialized novel distributing centrality more evenly than their in-volume contemporaries across the age of the feuilleton. Confirming this at the level of individual texts, once publication metadata become available, is the natural next step, as is extending the comparison to other national literatures with a comparable serial press. Corpus, annotations, and code are released under an open license (Appendix~\ref{appdx:code}).



\section*{Acknowledgements}
This study benefited from the PSL Research University's Major Research Program CultureLab, implemented by the ANR (reference ANR-10-IDEX-0001). Thierry Poibeau's research is partially funded by the French State, managed by the Agence Nationale de la Recherche, under the France 2030 program (grant reference ANR-23-IACL-0008) and by the European Union through the European Research Council (ERC, CANON, 101265457). Jean Barré's postdoctoral research is funded by the European Union through the European Research Council (ERC, MELT, 101221334). Views and opinions expressed are however those of the authors only and do not necessarily reflect those of the European Union or the European Research Council. Neither the European Union nor the granting authority can be held responsible for them.

\printbibliography

\appendix

\section{Robustness of the Trajectory} \label{appdx:robustness}
This appendix reports the controls summarized in Section~\ref{sec:res-ushape}, each of which addresses a way in which the U-shaped trajectory might be an artefact of measurement or corpus composition and not a property of the texts.

\paragraph{Character cutoff.} Since the networks retain the twenty most frequent characters per novel, we rebuild them with cutoffs of ten, thirty, and forty characters, which leaves both the shape and the position of the trough essentially unchanged (Table~\ref{tab:robustness}), so the trajectory is not an artefact of that threshold.

\begin{table}[ht]
  \centering
  \begin{tabular}{lrr}
    \toprule
    $N$ & vertex & $d_{@1900}$ \\
    \midrule
    10 & 1870 & 0.16 \\
    20 & 1868 & 0.19 \\
    30 & 1865 & 0.19 \\
    40 & 1862 & 0.18 \\
    \bottomrule
  \end{tabular}
  \caption{Top-$N$ robustness. Vertex year and effect size (Cohen's $d$ at the 1900 split) for the structural concentration metric (Gini of PageRank), for four values of the character cutoff. The vertex moves by only a few years under a four-fold change in cast size.}
  \label{tab:robustness}
\end{table}

\begin{figure}[ht]
  \centering
  \includegraphics[width=\linewidth]{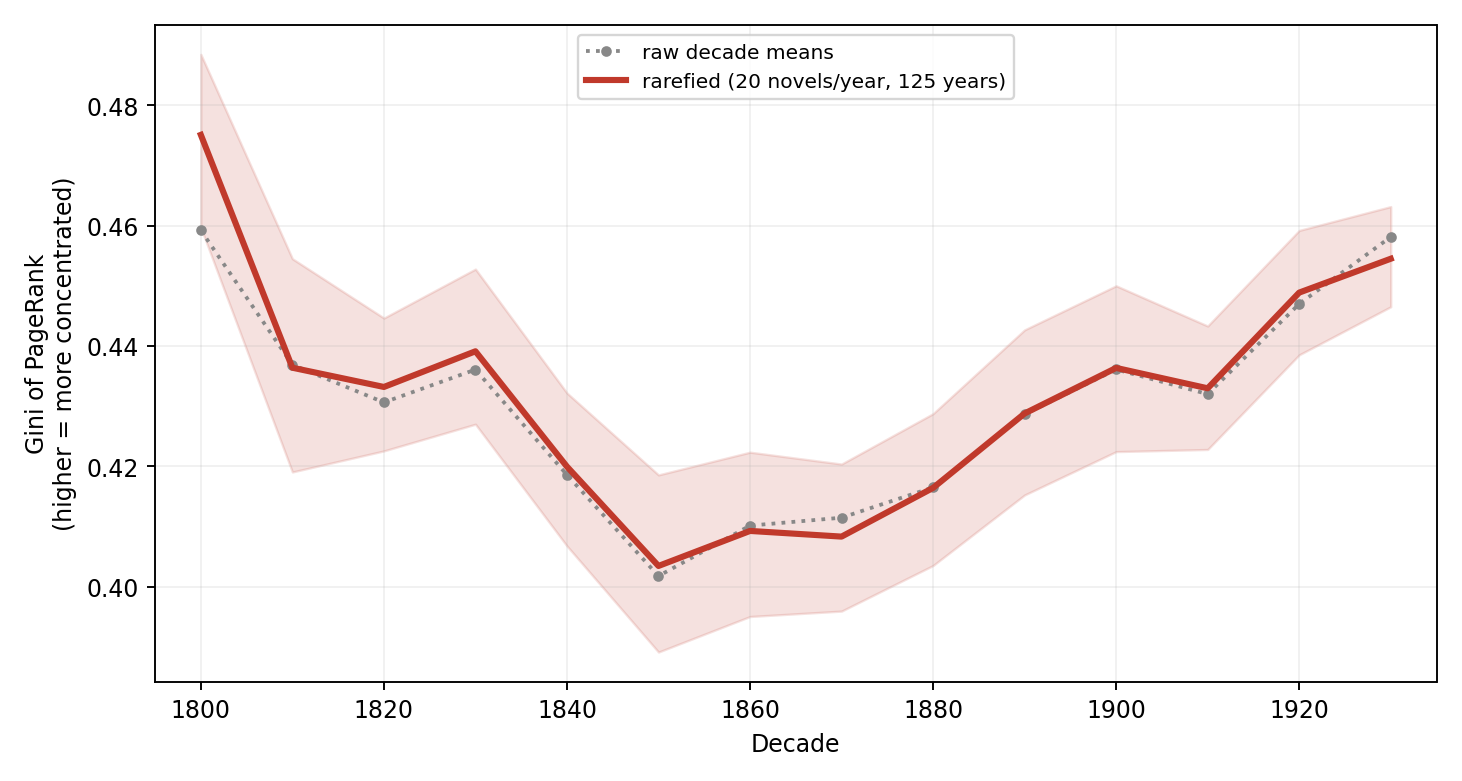}
  \caption{Density control by rarefaction. Raw decade means (grey, dotted) and the
  trajectory obtained by subsampling twenty novels per year (red, with 95\% band)
  coincide, showing that the U-shape is not produced by variation in the number of
  novels available per year.}
  \label{fig:rarefaction}
\end{figure}

\paragraph{Corpus density.} Because the corpus is unevenly populated over time, peaking in production between roughly 1850 and 1890, we subsample twenty novels per year (retaining the 125 years that reach that threshold) and average over 300 draws, to rule out that this uneven density shapes the trajectory. The U-shape and the position of its trough persist (Figure~\ref{fig:rarefaction}), with a rarefied vertex at 1869 (95\% interval 1865--1872). The series is cut at the end of the 1930s, because wartime publishing and copyright-driven gaps in the digitization leave the 1940s with 71 novels, and single years with as few as one, which no resampling can repair.

\paragraph{Author effects.} A diachronic trend could be an artefact of composition, with a few prolific authors and their idiosyncratic habits dominating particular decades. Author identity does account for a substantial share of the variance in concentration (an author fixed-effect model explains roughly a quarter of it, so writers do have recognizable ``character-system signatures''), but about three quarters of the variation is within-author, from one novel to the next. More important for the diachronic claim, the U-shape is not carried by individual authors, since adding author fixed effects to the period model leaves the period coefficients convex and slightly sharpens them, so the trough and rise persist even when each author is compared only to himself over time. Removing the ten most prolific authors leaves the trajectory essentially unchanged (vertex within two years, effect size at 1900 unchanged).

\section{Structure vs.\ Attention Concentration} \label{appdx:scatter}
\begin{figure}[ht]
  \centering
  \includegraphics[width=0.8\linewidth]{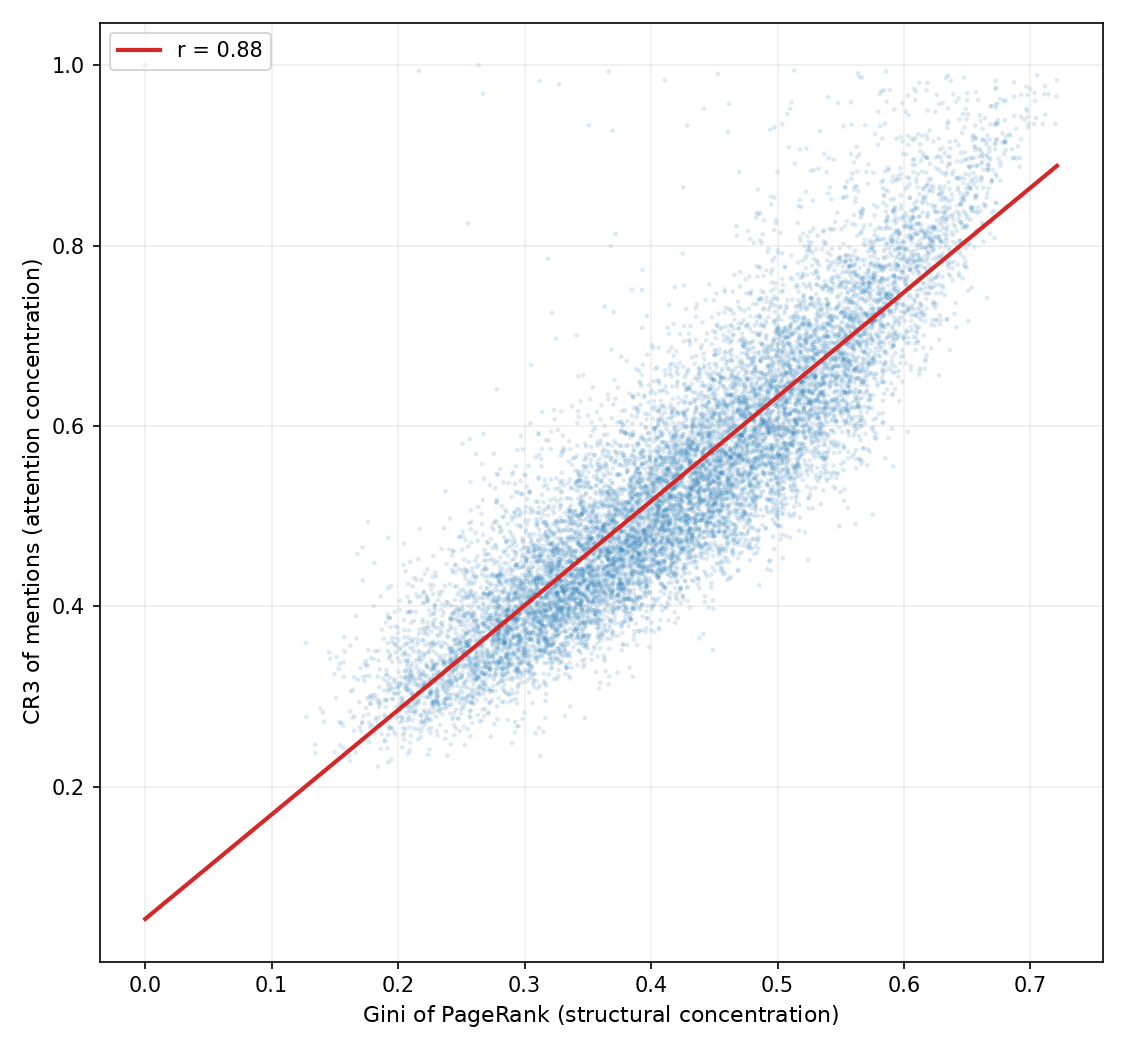}
  \caption{Structural concentration (Gini of PageRank, $x$) vs.\ attention concentration (CR3 of mentions, $y$) for every Gallica novel ($n=12{,}011$). Red line, linear fit ($r=0.88$).}
  \label{fig:scatter}
\end{figure}

Figure~\ref{fig:scatter} plots the structural concentration index (Gini of PageRank) against the attention index (CR3 of mentions) for every book in the Gallica corpus. The two are strongly correlated (Pearson $r=0.88$, Spearman $\rho=0.89$, $n=12{,}011$), a correlation that partly reflects their shared construction, since both are computed from the same extraction and the same twenty characters. The spread around the regression line corresponds to novels in which attention and structure decouple, such as a frequently named but structurally peripheral narrator, or a sparsely named but structurally pivotal connector, and these are the cases we target for close reading.

\newpage
\section{Character-Network Extraction, Extended Evaluation}
\label{appdx:failure_modes_appendix}

Table~\ref{tab:network_quality} provides the complete per-novel results of the cast identification and identity resolution evaluation described in Section~\ref{sec:failure_modes}, for the three fully annotated novels of the extended LitBank-Fr corpus and the nine novels with manually compiled character lists and sets of unique mentions. Results are expressed as percentages, with novels ordered by text length.

\begin{table*}[ht]
\centering
\small
\setlength{\tabcolsep}{5pt}
\renewcommand{\arraystretch}{1.15}

\resizebox{\textwidth}{!}{%
\begin{tabular}{rl l l rrr rrr}
\toprule
& & & & \multicolumn{3}{c}{\textbf{Cast identification}}
& \multicolumn{3}{c}{\textbf{Identity resolution}} \\
\cmidrule(lr){5-7} \cmidrule(lr){8-10}
Tokens & Year & Author & Title
& Precision & Recall & F$_1$
& Precision & Recall & F$_1$ \\
\midrule
26,127  & 1830 & Eugène Sue & Kernok le Pirate
& 100.00 & 100.00 & 100.00 & 95.64 & 84.27 & 89.60 \\

32,544  & 1843 & Honoré de Balzac & La Fille aux yeux d'or
& 100.00 & 62.50 & 76.92 & 95.57 & 93.22 & 94.38 \\

39,106  & 1886 & Charles Dickens & Le Grillon du foyer
& 100.00 & 66.67 & 80.00 & 95.76 & 90.37 & 92.99 \\

57,995  & 1882 & Jane Austen & Persuasion
& 100.00 & 94.12 & 96.97 & 92.66 & 89.09 & 90.84 \\

71,219  & 1731 & Abbé Prévost & Manon Lescaut
& 100.00 & 96.00 & 97.96 & 93.97 & 74.60 & 83.17 \\

95,672  & 1923 & Delly & Dans les ruines
& 100.00 & 100.00 & 100.00 & 95.48 & 76.99 & 85.25 \\

105,675 & 1908 & Gaston Leroux & Le Mystère de la chambre jaune
& 100.00 & 100.00 & 100.00 & 93.04 & 82.03 & 87.19 \\

109,324 & 1843 & Honoré de Balzac & Le Père Goriot
& 100.00 & 76.47 & 86.67 & 97.67 & 78.74 & 87.19 \\

112,221 & 1832 & George Sand & Indiana
& 100.00 & 79.17 & 88.37 & 94.18 & 73.88 & 82.81 \\

152,347 & 1866 & Émile Gaboriau & L'Affaire Lerouge
& 100.00 & 100.00 & 100.00 & 95.55 & 77.10 & 85.34 \\

197,046 & 1869 & Gustave Flaubert & L'Éducation sentimentale
& 100.00 & 100.00 & 100.00 & 94.54 & 85.35 & 89.71 \\

284,497 & 1844 & Alexandre Dumas & Les Trois Mousquetaires
& 100.00 & 100.00 & 100.00 & 97.79 & 97.41 & 97.60 \\
\bottomrule
\end{tabular}%
}
\caption{Cast identification and identity resolution evaluation across the twelve novels. Precision, recall and F$_1$ are expressed as percentages.}
\label{tab:network_quality}
\end{table*}

\section{Specification Curve for the Serialization Test} \label{appdx:specs}
\begin{figure}[ht]
  \centering
  \includegraphics[width=\linewidth]{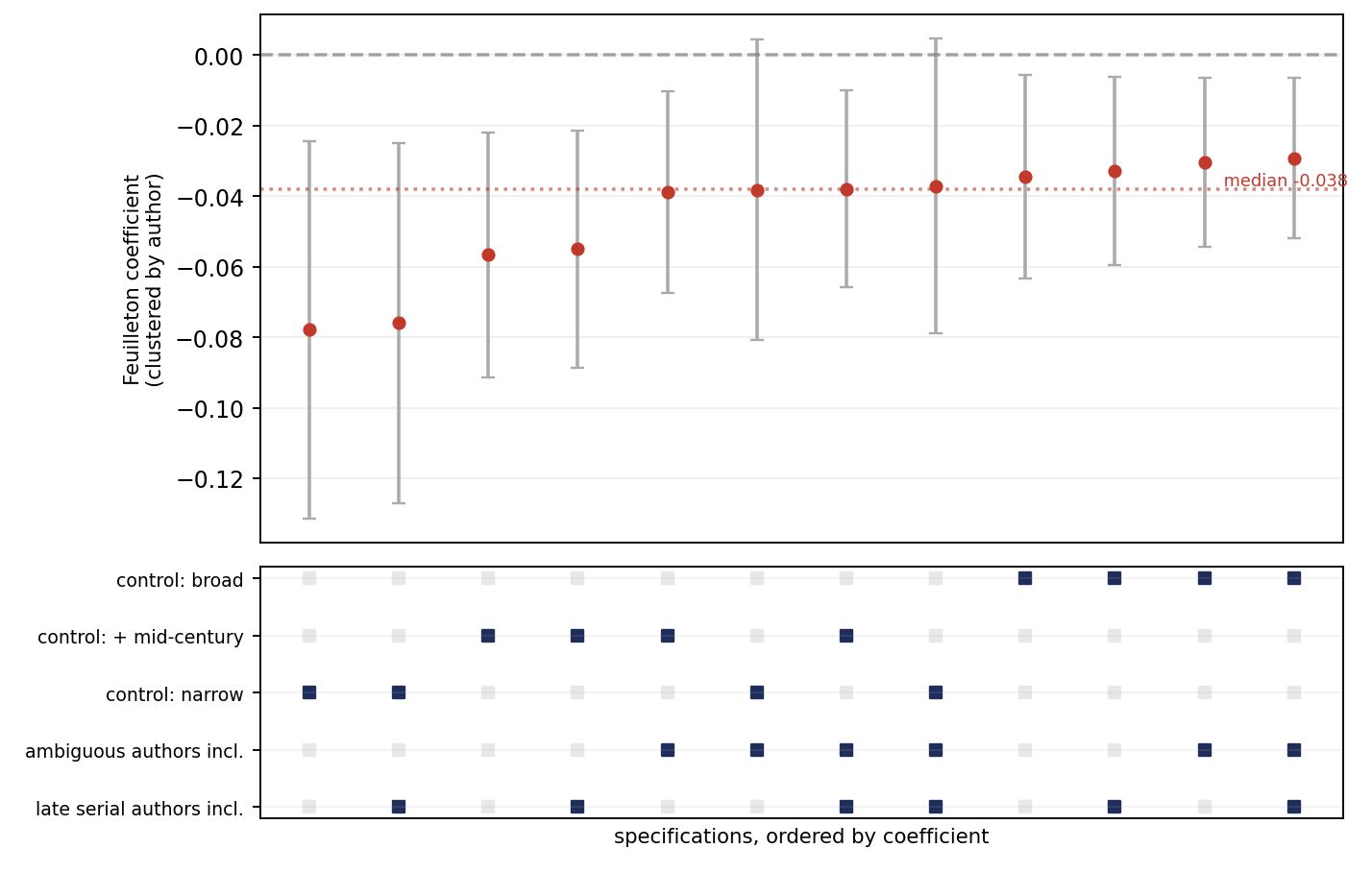}
  \caption{Specification curve for the serialization test. The upper panel shows the feuilleton coefficient with its author-clustered 95\% confidence interval under each of twelve compositions of the two author groups, ordered by magnitude, with the dotted line marking the median. The lower panel shows which options are active in each specification (dark squares).}
  \label{fig:specs}
\end{figure}

The feuilleton and control groups are defined by editorial history, but their exact composition involves judgement about how broadly to draw the literary control, whether to retain the authors who worked in both circuits, and whether to include the serial writers of the 1890s and 1900s. We cross these three choices and estimate the model under all twelve resulting compositions (Figure~\ref{fig:specs}).

With standard errors clustered by author, the coefficient is negative in every specification, ranging from $-0.029$ to $-0.078$ (median $-0.038$), and significant at the 5\% level in ten of the twelve. The two exceptions are the specifications built on the narrowest control (185 novels), whose intervals are correspondingly wide, and no specification returns a positive estimate. Two regularities run through that spread, as excluding the authors who wrote in both circuits always \emph{strengthens} the estimate, which one would predict if they genuinely straddle the two regimes, and broadening the control always moves it toward zero while tightening the interval, as a more heterogeneous and hence more conservative comparison would. Interval width depends almost entirely on the breadth of the control (0.067 at its narrowest, 0.022 at its broadest), and much less on the treatment of ambiguous authors.

\paragraph{Recurring-hero series.} Within the labelled group of 35 serialization-oriented authors, to which the comparison of Section~\ref{sec:res-feuilleton} is internal, the 7 who built series around a returning figure are less concentrated than the other 28 ($d=-1.78$, $p<10^{-4}$). Admitting the four series authors excluded from the main test as publication-ambiguous (Ponson du Terrail, Leblanc, Leroux, Sazie), among them the creators of Rocambole and Arsène Lupin, changes the size of the gap but not its direction, with 11 authors against 28, $d=-0.59$, $p=0.03$. On either population, the recurring hero is associated with a \emph{more} distributed character-system.

\section{Concentration Metric Definitions} \label{appdx:metric}
Let $x_1 \ge x_2 \ge \dots \ge x_n$ be the values of interest for the $n$ retained characters of a novel (mention counts for the attention layer, PageRank scores for the structural layer) and let $\bar{x}$ be their mean.

The \emph{Gini coefficient} is
\[
G \;=\; \frac{\sum_{i=1}^{n}\sum_{j=1}^{n} |x_i - x_j|}{2n^{2}\bar{x}},
\]
taking the value 0 when all characters carry equal weight and approaching 1 as a single character absorbs the whole distribution.

The \emph{concentration ratio} of the top three characters is
\[
\mathrm{CR3} \;=\; \frac{x_1 + x_2 + x_3}{\sum_{i=1}^{n} x_i},
\]
the share of the total held by the three largest. Both indices are computed at the operating cutoff $n = N = 20$.

\section{Concentration by Genre} \label{appdx:strata}
Table~\ref{tab:genres} reports mean concentration by genre in the auxiliary corpus, whose records carry a finer generic classification than the Gallica metadata, restricted to 1800--1950 and to genres with at least twenty novels.

\begin{table}[ht]
  \centering\small
  \begin{tabular}{lrrr}
    \toprule
    Genre & $n$ & mean & vs.\ corpus \\
    \midrule
    short stories        & 175 & 0.346 & $-0.070$ \\
    adventure            & 348 & 0.388 & $-0.028$ \\
    detective            & 166 & 0.404 & $-0.013$ \\
    historical novel     & 145 & 0.411 & $-0.005$ \\
    travel narrative     &  30 & 0.413 & $-0.003$ \\
    memoir, autobiography&  54 & 0.436 & $+0.019$ \\
    fantastic            &  56 & 0.442 & $+0.025$ \\
    juvenile fiction     & 134 & 0.458 & $+0.042$ \\
    cycles and series    & 162 & 0.463 & $+0.046$ \\
    epistolary           &  31 & 0.487 & $+0.071$ \\
    sentimental novel    &  74 & 0.498 & $+0.082$ \\
    \bottomrule
  \end{tabular}
  \caption{Concentration by genre (auxiliary corpus, 1800--1950, corpus mean 0.416). Forms built on a collective or serially renewed cast occupy the lower end, and forms organized around a personal plot the upper end.}
  \label{tab:genres}
\end{table}

\section{Code and Data Availability} \label{appdx:code}

The metadata, scripts and output data used in this study are available in a
GitHub repository at \url{https://github.com/lattice-8094/character_concentration/}. The curated \emph{Hors d'usage} corpus is archived on Zenodo at \url{https://doi.org/10.5281/zenodo.20233178}.


\end{document}